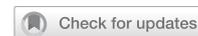

# Response to: Significance and stability of deep learning-based identification of subtypes within major psychiatric disorders. *Molecular Psychiatry* (2022)

**TO THE EDITOR:**
Recently, Winter and Hahn [1] commented on our work on identifying subtypes of major psychiatry disorders (MPDs) based on neurobiological features using machine learning [2]. They questioned the generalizability of our methods and the statistical significance, stability, and overfitting of the results, and proposed a pipeline for disease subtyping. We appreciate their earnest consideration of our work, however, we need to point out their misconceptions of basic machine-learning concepts and delineate some key issues involved.

**Subtyping diseases**
Subtyping diseases, such as MPDs in [2], is *a task of clustering a set of unlabeled data (i.e., patients) with no definition of target subtypes and no known number of subtypes*. Clustering is fundamentally different from classification where training data with class labels are provided.

When no subtype/class label is available, most concepts and techniques developed for classification do not apply. However, Winter and Hahn proposed a clustering pipeline [1] consisting of components primarily applicable to classification, including generalization, statistical significance test, overfitting avoidance, and cross-validation.

**Generalization and overfitting**
Generalization is a concept for classification (supervised learning). In classification, a model is learned using the training data with class labels available to the learning process, the model is then applied to predict labels of the test data, and the predicted and true class labels are compared to measure the generalizability of the learned model on unseen samples. Clustering is completely different—no class label is provided, and no data distribution is known. As such, the concepts of generalization and overfitting do not apply to clustering.

Winter and Hahn commented on the generalizability and data overfitting of our methods. They suggested that "to assess the generalizability of the autoencoder data representation, a cross-validation framework should be used dividing the sample into training and test set" and "the amount of overfitting in small samples is generally estimated through cross-validation" [1]. These statements highlight their misunderstanding of machine-learning techniques. First, for our subtyping problem, no class/cluster label was available. Second, cross-validation is a technique for measuring the generalizability of a classification method against overfitting the training data with class labels [3]. Since generalizability does not apply to clustering methods, as discussed above, cross-validation is irrelevant as well. Third, overfitting is difficult to assess in clustering methods because there are no class/cluster labels to quantify clustering results. Our use of autoencoder was to generate more perturbed data to increase the robustness of the results, which will be discussed shortly.

**Cluster validity**
A central question for clustering is how to evaluate the validity of clustering results. Unfortunately, there is no universal solution due to the lack of class labels in clustering. With no class labels as a reference, a countless number of possible cluster structures may exist. "A cluster is a subjective entity that is in the eye of the beholder and whose significance and interpretation requires domain knowledge" [4]. It is challenging to define the "best" cluster structure; worse, no best cluster structure may exist at all. "Clustering should not be treated as an application-independent mathematical problem, but should always be studied in the context of its end-use" [5].

Any dataset can undergo clustering, and the meaning of clustering results is up to the end-user for interpretation. A clustering algorithm is developed based on hypothesis of what constitutes a meaningful cluster structure as envisioned by the end-user. Such a hypothesis is characterized by an objective function or score to be optimized [5, 6]. Different hypotheses are pillared by different assumptions regarding envisioned cluster structures, giving rise to different objective functions and producing distinct results. Therefore, no standard mechanism exists to compare clustering methods [5]. Clustering is merely a data exploration apparatus and its value depends on the ultimate application where clustering results are incorporated.

Nevertheless, it is important to validate clustering results by additional data with orthogonal modalities (i.e., from different data sources) and by the end applications. In our study, we identified new subtypes for MPDs based on brain neuroimaging patterns (from functional Magnetic Resonance Imaging (fMRI) data). The results were assessed by cortical thickness and white matter integrity (from Diffusion Tensor Imaging data), polygenic risk scores (from genome-wide genotyping data), and tissue profile for risk gene expression (from gene-expression profiling data). We also examined the effects of medication on symptom severity (from clinical data) to elucidate pharmacologic effects within each subtype. These validation analyses showed significant alignment between the subtyping results and these other biological features [2].

**Statistical significance**
Winter and Hahn suggested evaluating result significance by statistical tests against a statistical null model to ensure the result clusters do exist. Defining a null model is equivalent to asking what typical clusters or distributions the given data should have. Much effort has been devoted to developing null models for clustering, as discussed in the whole Chapter 4 of the book by Bezdek [7]. It has become evident that effort to define a null model is futile because there is no universal definition of a "cluster" and "the concept of no structure in the dataset (one possible null hypothesis) is far from clear" [6].

Winter and Hann suggested adopting the procedure proposed by Liu et al. [8] to estimate an empirical data distribution as a null model [1]. Unfortunately, to make this procedure work, "we define a cluster as data coming from a single Gaussian distribution" [8]. Therefore, they suggested injecting an unjustified data distribution into the analysis. Such distribution can only be taken as a hypothesis on the data and would itself need to be tested. In our study, the data distribution and valid null model are unknown and cannot be defined. Furthermore, statistical tests based on a hypothesized Gaussian distribution cannot ensure the validity of a subtyping result.

### Stability and robustness

Stability and robustness are key properties for clustering, as pointed out by Winter and Hahn [1]. Our methods were designed to enhance these properties. In assessing the stability and robustness of a clustering result, perturbed versions of the given data can be used. Perturbed data can be generated by either random sampling of the original data or random projection of the data to low-dimensional spaces [9]. We chose the second approach and adopted autoencoder to perform multiple projections. Such perturbed data were used to extract latent features in the data to facilitate the identification of robust clusters. We did not employ the first approach because sampling from a small set of high-dimensional data can easily fail to capture latent patterns in the data.

### Concluding remarks

Our primary objective was to identify subtypes of MPDs for understanding functional brain patterns to better treat these heterogeneous diseases [2]. Our study was not for the classification of MPDs. As such, techniques developed for classification do not apply to our study. Clustering is an approach for data exploration and the utility and value of clustering results are application dependent. There is no universal or standard definition of valid clusters. It is critically important to assess and validate clustering results using additional data from different sources. In our study, we showed significant alignment of the MPD subtype results from fMRI data with the biological features in data from four other sources, offering support for the significance of the identified MPD subtypes, which should be replicated in further clinical studies.

Xizhe Zhang[1,2]✉, Fei Wang [ID][1]✉ and Weixiong Zhang [ID][3,4]✉
[1]Early Intervention Unit, Department of Psychiatry, Affiliated Nanjing Brain Hospital, Nanjing Medical University, Nanjing, China. [2]School of Biomedical Engineering and Informatics, Nanjing Medical University, Nanjing, China. [3]Department of Health Technology and Informatics, The Hong Kong Polytechnic University, Hong Kong, Hong Kong. [4]Department of Computer Science and Engineering, Department of Genetics, Washington University, St. Louis, MO, USA.
✉email: zhangxizhe@njmu.edu.cn; fei.wang@yale.edu; weixiong.zhang@polyu.edu.hk

### ACKNOWLEDGEMENTS
The authors were supported by research grants from the National Natural Science Foundation of China Funding support: National Natural Science Foundation of China (62176129 to XZ), Natural Science Foundation for Distinguished Young Scholars (81725005 to FW).

### AUTHOR CONTRIBUTIONS
XZ, FW, and WZ wrote the paper.

### COMPETING INTERESTS
The authors declare no competing interests.

### ADDITIONAL INFORMATION
**Correspondence** and requests for materials should be addressed to Xizhe Zhang, Fei Wang or Weixiong Zhang.

**Reprints and permission information** is available at http://www.nature.com/reprints

**Publisher's note** Springer Nature remains neutral with regard to jurisdictional claims in published maps and institutional affiliations.